\documentclass{article}
\usepackage{ijcai26}

\usepackage{times}
\usepackage{soul}
\usepackage{url}
\usepackage[hidelinks]{hyperref}
\usepackage[utf8]{inputenc}
\usepackage{graphicx}
\usepackage{algorithm}
\usepackage{algpseudocode}
\usepackage{amsmath}
\usepackage{amsthm}
\usepackage{booktabs}
\usepackage[switch]{lineno}
\usepackage{epstopdf}
\usepackage{subcaption}
\usepackage{array}
\usepackage{cleveref}
\usepackage{enumitem} 
\crefname{figure}{Figure}{Figures}
\Crefname{figure}{Figure}{Figures}
\DeclareCaptionFont{ninept}{\fontsize{9}{10.8}\selectfont}
\title{Hybrid-Adaptive Thread Tuning to Mitigate Simulation Execution Bottlenecks in High-Performance Reinforcement Learning Inference}

\author{
Jiming Su$^1$
\and
Hantao Hua$^{2}$\and
Lujia Yin$^{1}$\and
Yiping Yao$^{1}$\And
Feng Zhu$^1$\footnotemark[1]\\
\affiliations
{\normalfont\fontsize{12}{14}\selectfont $^1$College of Systems Engineering, National University of Defense Technology}\\
{\normalfont\fontsize{12}{14}\selectfont $^2$College of Computer Science and Technology, National University of Defense Technology}\\
\emails
\{jimings, ht\_hua, yinlujia10, ypyao, zhufeng\}@nudt.edu.cn
}

\begin{document}

\maketitle

\begin{abstract}
    In simulation-in-the-loop decision-making systems, reinforcement learning (RL) inference is often constrained by simulator-side execution overhead, where workloads are highly dynamic and sensitive to runtime thread configurations. Existing multithreaded strategies struggle to match thread resources before or during execution, causing resource contention, scheduling overhead, and reduced throughput. Through empirical analysis, we identify the ratio of task execution time to scheduling time as the key factor determining the optimal thread count. Building on this insight, we propose AutoThread, a hybrid adaptive thread-tuning method for mitigating simulation bottlenecks in RL inference. AutoThread employs a Physics-Informed Neural Operator (PINO) as a thread-count predictor and incorporates a finite-source M/M/1 queueing model to constrain and guide prediction, enabling fast and accurate estimation under dynamic workloads. It further performs load-aware online fine-tuning to compensate for prediction errors and refine resource allocation. Experiments show that AutoThread improves average speedup by 18.4\% over static strategies, achieves average throughput of 1.7× and 1.8× that of XGBoost and Reinforcer, respectively, and reduces execution time by up to 83.8\% compared with state-of-the-art methods. Our code and dataset are publicly available at \url{https://github.com/suchenjm/AutoThread}.
\end{abstract}

\section{Introduction}
As artificial intelligence is increasingly used for decision-making in complex dynamic systems, reinforcement learning with simulation-in-the-loop (SiL-RL) has become a core paradigm for wireless communication network optimization and swarm intelligence control\cite{feriani2021single,hu2024review}. Its operational efficiency, however, directly limits real-time decision-making. SiL-RL performance depends on three components: simulation-side model computation (e.g., social network dynamic evolution models\cite{wu2022mixed}), agent-side algorithm inference (e.g., Deep Q-Network decision-making\cite{zhang2021parallel}), and their communication/synchronization. As simulation models grow increasingly complex (e.g., high-precision channel interference models\cite{eldeeb2025offline} in communication networks or high-fidelity aerodynamic models\cite{riboldi2025formation} in UAV swarms), the simulation side becomes the critical bottleneck\cite{morgado2025evaluating}, increasing end-to-end latency and impairing real-time decisions. This paper therefore accelerates simulation-side performance to break this bottleneck and support efficient SiL-RL systems.

\begin{figure}
    \centering
    \includegraphics[width=1\linewidth]{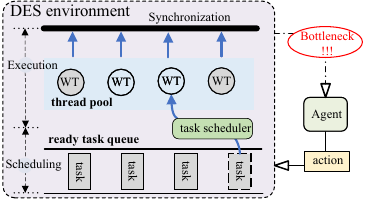}
    \caption{Diagram of the DES environment execution process}
    \label{fig:figure0}
\end{figure}

Existing SiL-RL simulation environments typically adopt Discrete Event Simulation (DES) and use multithreading on multicore processors for parallel acceleration\cite{belsare2022reinforcement}. They mainly employ two architectures: centralized queues with thread pools and distributed queues with work stealing\cite{risco2022unified}. The centralized queue, with dynamic load balancing and lower synchronization overhead, is well suited to SiL-RL inference, where agent activation rates and task loads fluctuate across simulation cycles\cite{lin2025hap}. This work therefore focuses on simulation acceleration under the centralized queue architecture.

However, centralized queues are highly sensitive to the number of worker threads (WTs): all WTs share one task queue, and task acquisition and distribution are handled serially by the scheduler\cite{samadi2024time}. Thus, the scheduler may become a serial bottleneck: as WTs increase, scheduler contention, lock waiting, and busy waiting rise sharply, negating or even reversing parallel gains. Selecting WTs according to workload characteristics is therefore key to improving simulation-side performance\cite{li2025adwtune}. First, the optimal WT count depends on workload intensity and hardware factors such as core frequency, memory bandwidth, and cache hierarchy, which vary across platforms and are hard to know a priori. Second, SiL-RL workloads are highly dynamic; agent activation rates, task execution times, and interaction intensities fluctuate across simulation stages, making fixed WT counts suboptimal. Runtime noise and external disturbances further obscure transient interference versus persistent workload shifts. Together, these factors prevent precise WT allocation before or during runtime.

To address this, we propose AutoThread, a hybrid adaptive thread-tuning method for SiL-RL simulation acceleration. The core idea is to construct a WT count predictor where a neural network is guided and constrained by a theoretical performance model. By introducing a dynamic load-aware adaptive fine-tuning strategy, we achieve significant improvements in simulation performance, thereby supporting high-performance inference in RL decision-making systems. The main contributions of this paper are as follows:

\begin{itemize}
\item We open-source a large-scale multithreaded trajectory dataset for DES environment. Through empirical analysis, we derive three findings regarding the relationship between WT count and simulation performance.
\item We develop a Physics-Informed Neural Operator (PINO) thread predictor, utilizing a queueing model to constrain and guide the training process.
\item We propose AutoThread, which leverages PINO-based thread prediction and load-aware online fine-tuning to achieve fast and adaptive WT count allocation.
\end{itemize}

\section{Empirical Study}

We first introduce the dataset, then empirically reveal the relationship between WT count and execution performance to guide AutoThread's model construction and tuning strategy.

\subsection{Dataset}
We constructed the multithreaded trajectory dataset for the DES environment by collecting over 10,000 multithreaded execution samples across both Intel and AMD CPU hardware platforms. The dataset includes two application cases:

{\bfseries PCS:} The Personal Communication System(PCS) \cite{carothers2002ross} is a benchmark DES environment that simulates wireless communication processes within a network topology and is widely used to test decision algorithms in communication networks. In this application, each agent represents a communication node, continuously generating five types of communication tasks.

{\bfseries UAV:} This is a case study based on the DES paradigm, simulating the autonomous navigation and obstacle avoidance of multiple unmanned aerial vehicles (UAVs) within a specific area. The paths are derived from real-world flight data\cite{flightaware_uk_2025_online}. Each agent represents a UAV, involving both perception and navigation tasks.

To emulate high-fidelity simulation, we embedded array dot products and iterative power computations into each task, producing execution times between $10\,\mu\text{s}$ and $800\,\mu\text{s}$. With profiling tools, we captured each sample's execution trajectories under various WT counts. The data include hardware utilization, application logs, and micro-architectural features, totaling 34 variables. To our knowledge, this is the first public DES trajectory dataset for RL systems and among the largest single-node thread-level trace sets; we release it to support studies on multithreaded behavior and DES acceleration.

\subsection{Empirical Observations}

We conduct three experiments to characterize how workload, hardware, and micro-architectural metrics jointly affect performance. Our analysis yields the following observations.

\subsubsection{Dependence of Optimal WTs on Workload Intensity}

\begin{figure}[t]
    \centering
    \includegraphics[width=\linewidth]{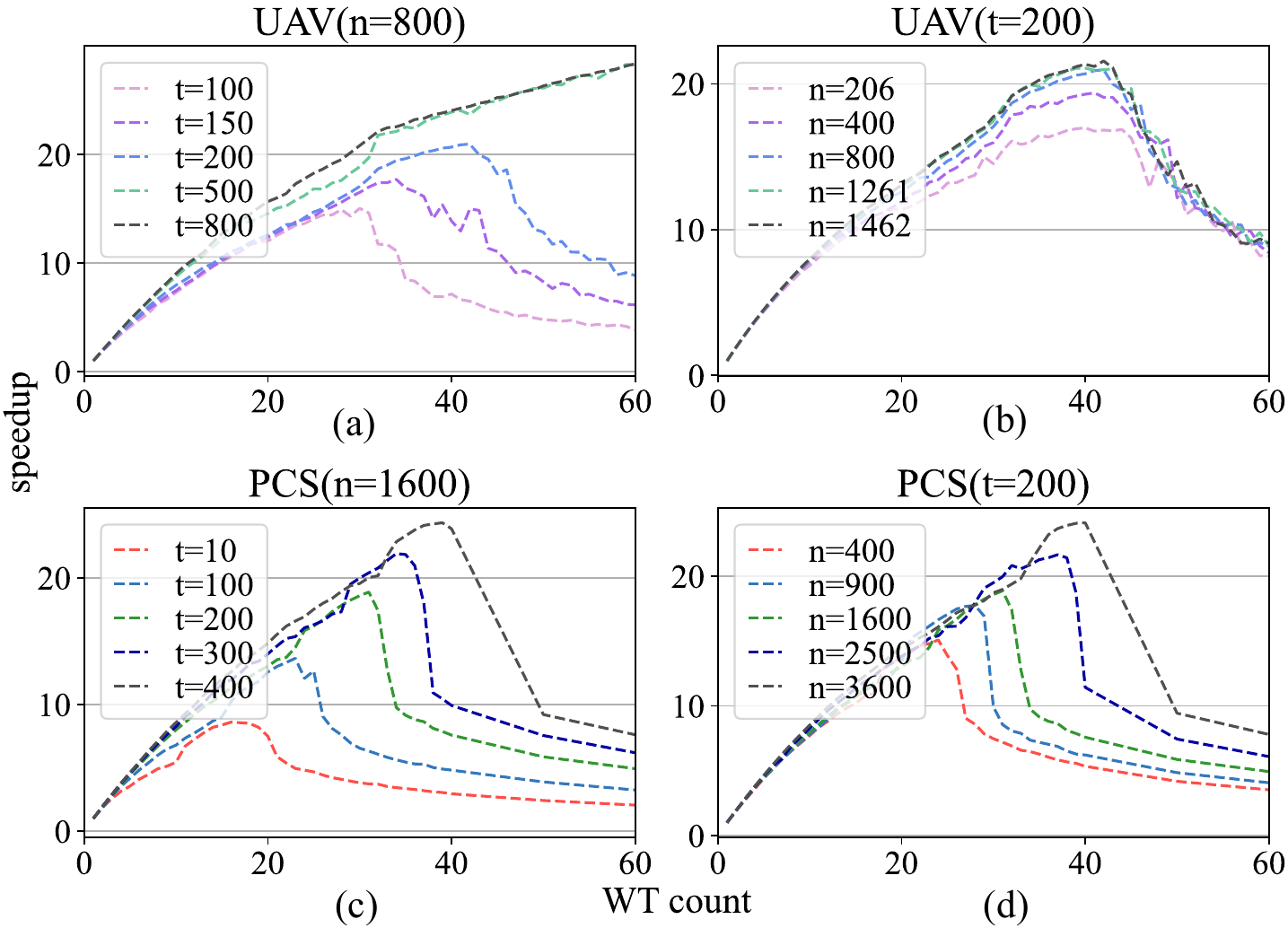}
    \caption{Variation of speedup with WT count under different different application parameters, where $t$ denotes the average task execution time and $n$ denotes the number of agents}
    \label{fig:fig7}
\end{figure}

\begin{figure}[t]
    \centering
    \includegraphics[width=0.85\linewidth]{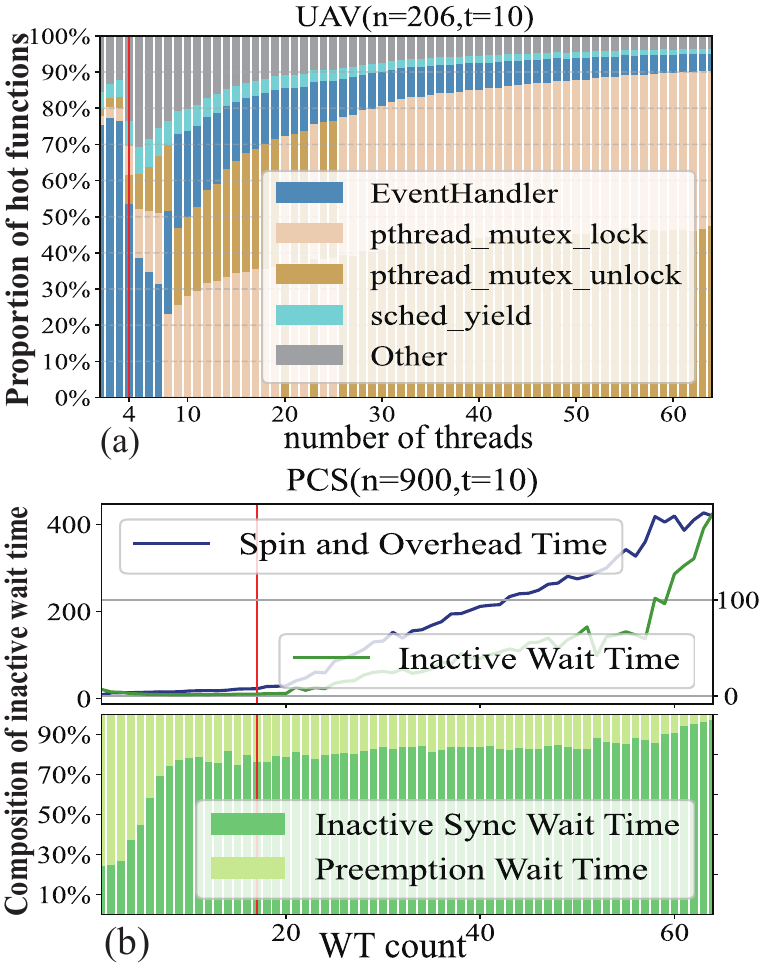}
    \caption{Hotspot function distribution and thread time consumption in DES environment}
    \label{fig:fig6}
\end{figure}

Figure~\ref{fig:fig7} shows speedup versus WT count under different agent populations and task execution times. In UAV, each agent always generates two fixed tasks with execution time $t$; in PCS, each agent generates a variable number of tasks whose execution times follow a Poisson distribution with mean $t$. The optimal WT count is positively correlated with average task execution time: higher workload intensity increases the optimal WT count because longer tasks enlarge parallelizable regions and let more threads improve performance. Conversely, with fixed execution time, the optimal WT count remains nearly unchanged regardless of agent scale(Figure~\ref{fig:fig7}(b)). The variance in Figure~\ref{fig:fig7}(d) arises because changing the number of PCS agents alters the task-queue execution-time distribution and thus its mean.

    

\subsubsection{Contention-Induced Waiting as the Cause of Degradation}
We conduct detailed analysis on a subset of samples using VTune. Figure~\ref{fig:fig6} displays the distribution of hotspot functions and thread time consumption. As shown in Figure~\ref{fig:fig6}(a), the proportion of time spent on lock operations rises rapidly as the WT count increases. Once the lock-wait time exceeds the task execution time, system performance begins to degrade. Figure~\ref{fig:fig6}(b) further shows that the critical point at which busy waiting starts to increase sharply around the optimal WT count is insensitive to inactive wait times such as inactive synchronization and preemption.

\begin{figure}
    \centering
    \includegraphics[width=1\linewidth]{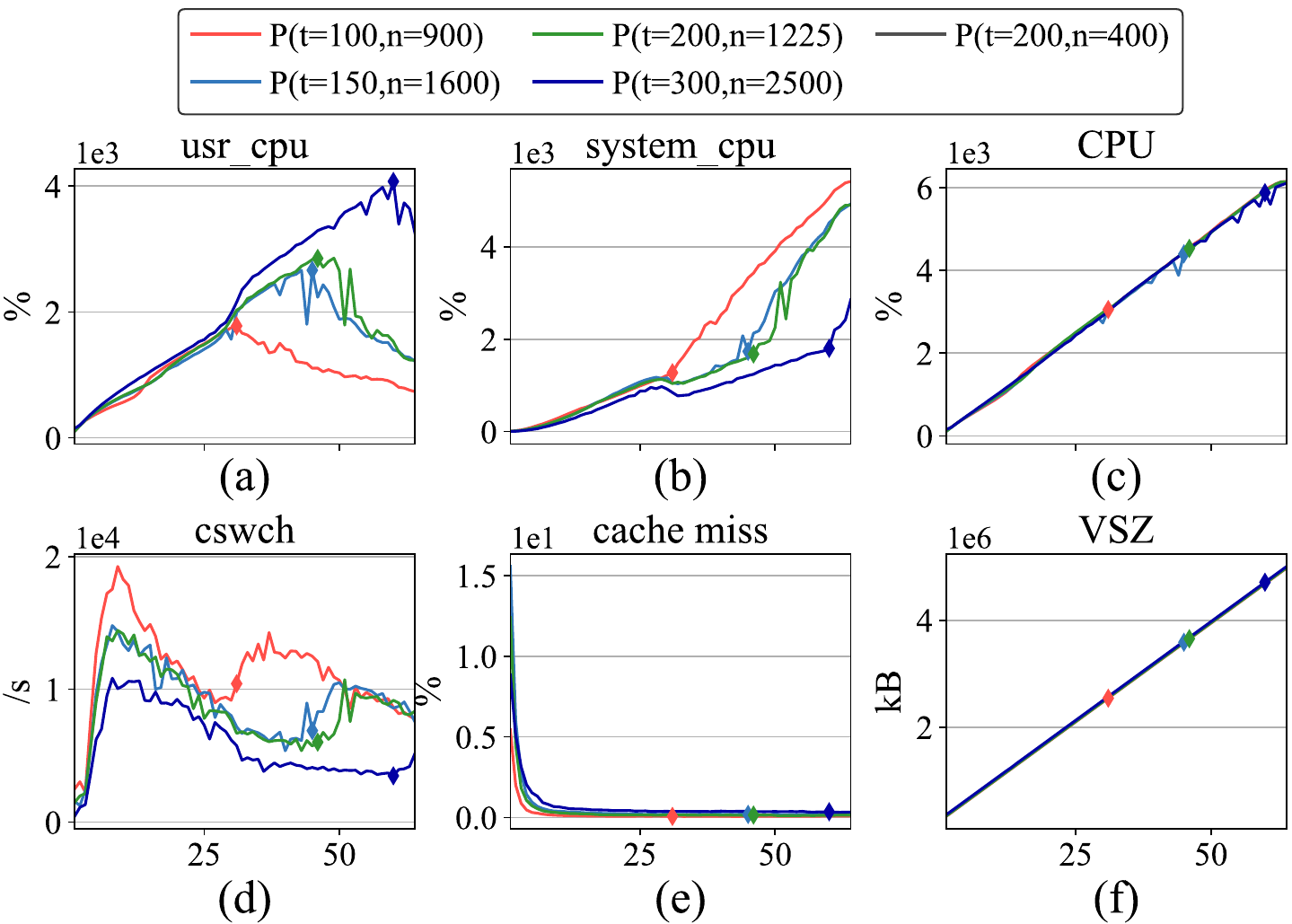}
    \caption{Variation of DES environment's micro-architectural Metrics with WT count}
    \label{fig:figure2}
\end{figure}

Although busy waiting time is key to the optimal WT count, measuring it online is challenging\cite{jancauskas2019predicting}. We therefore use lightweight metrics to indirectly reflect busy-waiting status, as shown in Fig \ref{fig:figure2}. Figure~\ref{fig:figure2}(a) shows that $usr\_cpu$ peaks near the optimal WT count, where CPU cycles are most efficiently used for parallel tasks. Once the WT count exceeds the optimum, $system\_cpu$ (Figure~\ref{fig:figure2}(b)) rises sharply, indicating a shift from effective computation to OS scheduling and thread waiting. Since busy-waiting threads still occupy physical cores, aggregate CPU utilization does not accurately reflect the optimal WT count (see Figure~\ref{fig:figure2}(c)). Figure~\ref{fig:figure2}(d) further shows that the context switch rate reaches a local minimum near the optimum, suggesting that an appropriate thread count reduces unnecessary scheduling switches. In contrast, cache miss rate (Figure~\ref{fig:figure2}(e)) and memory consumption (Figure~\ref{fig:figure2}(f)) affect DES performance but not optimal WT selection.



\subsubsection{Linear Interference Under Multicore Concurrency}
\begin{figure}[t]
    \centering
    \includegraphics[width=1\linewidth]{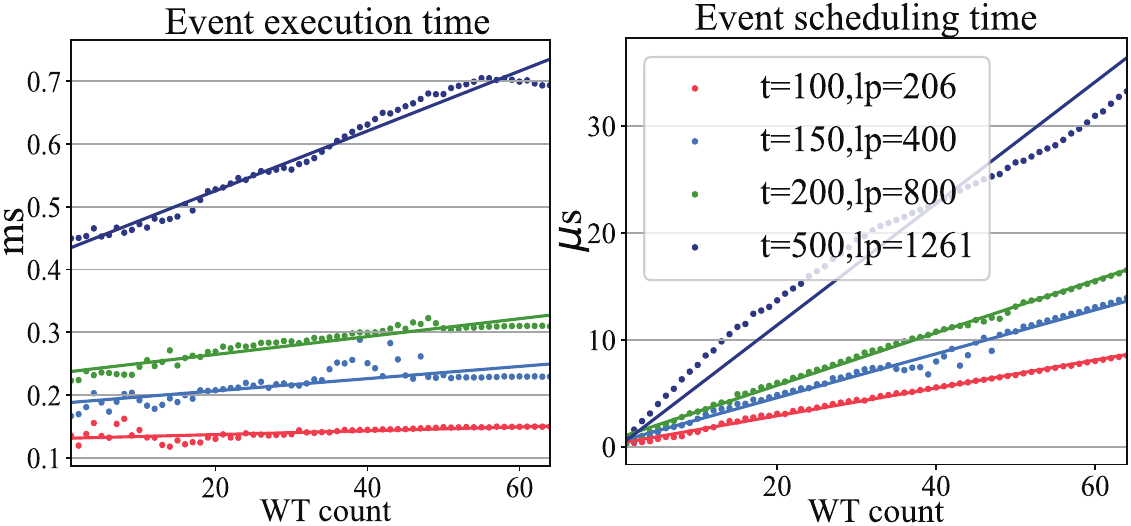}
    \caption{Performance interference among multiple WTs}
    \label{fig:figure5}
\end{figure}

Figure~\ref{fig:figure5} records the task execution time and task scheduling time for an individual working thread under varying numbers of active threads. The results show that as the number of active threads increases, both the execution and scheduling times for a single thread grow linearly. This indicates that the performance degradation of individual threads in a multithreaded environment can be approximated by a linear model, which captures the aggregate pressure exerted by all threads on a single thread's runtime. The slope and intercept of these linear functions depend on hardware characteristics (e.g., core frequency, cache size) and workload intensity.

\section{Methodology}
Based on the empirical findings in Section 2, we propose AutoThread. AutoThread consists of two core components: a PINO predictor and a dynamic tuner. This approach adopts a two-stage optimization strategy of online prediction followed by dynamic adjustment: during the online prediction stage, it rapidly estimates the optimal initial thread count based on the characteristics of the current workload; during the dynamic adjustment stage, it performs fine-grained adaptive thread tuning according to runtime feedback.

\begin{figure*}
    \centering
    \includegraphics[width=01\linewidth]{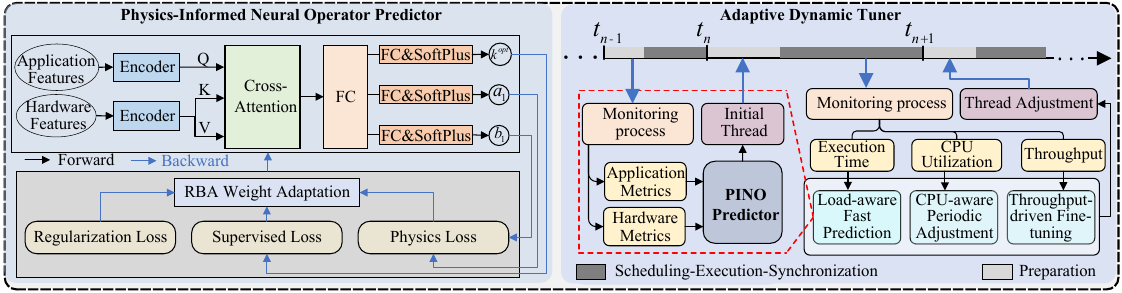}
    \caption{ Overall workflow of AutoThread}
    \label{fig:AutoThread}
\end{figure*}

\subsection{Model-Informed PINO Predictor}

\paragraph{Performance Model.}In typical SiL-RL inference workflows, the simulator must process a large number of computation tasks generated by multiple agents within each simulation step. We decompose the execution of a single simulation cycle into three sequential phases: task execution, task scheduling, and synchronization. WT reconfiguration is performed at the granularity of simulation cycles, during which the WT count remains fixed.



During the task execution phase, WTs process assigned tasks in parallel. Since the system must wait for all threads to finish before synchronization, the total execution time is determined by the slowest thread:
\begin{equation}
 T_{exe}(k)=\left\lceil \frac{N}{k} \right\rceil*\frac{1}{\mu_e}
\nonumber
\end{equation}
where $N$ is the total number of tasks within the current simulation step; $k$ is the number of WTs; and $\mu_e$ denotes the average service rate of tasks within the current simulation step.

Scheduling is inherently serial because the shared scheduler serves at most one WT at a time. We model the scheduler as an M/M/1 server and WTs as customers in a finite-source closed queueing system, where a WT requests scheduling only after finishing its current task, yielding an effective arrival rate $\lambda=k\mu_e$. When the system is stable, steady-state Continuous-Time Markov Chain (CTMC) analysis gives:



\[
\begin{aligned}
  T_{sch}=N\ast(\frac{k}{\mu_b(1-P_0)}-\frac{1}{k\mu_e}), &
  P_0=\frac{1}{\sum_{i=0}^{k}{\frac{k!}{(k-i)!}{(\frac{k\mu_e}{\mu_b})}^i}}
\end{aligned}
\]
where  $\mu_b$ denotes the scheduler service rate and $P_o$ is the idle probability of the scheduler. When $k\mu_e$ exceeds $\mu_b$, the scheduler becomes the dominant bottleneck and the total time is bounded by the scheduler capacity:
\[
T_{exe}(k)+T_{sch}(k)=\frac{N}{\mu_b}+\frac{1}{\mu_e}
\]
This captures the saturation regime under oversubscription.

The synchronization phase (e.g., global state update and clock advance by the master thread) is treated as constant with respect to $k$ when focusing on WT selection.

Based on Section 2.2.3, resource contention causes performance degradation that grows linearly with active WT count:
\[
\begin{aligned}
\mu_b(k) &= \frac{1}{a_1k+t_{base1}},&
\mu_e(k) &= \frac{1}{a_2k+t_{base2}}
\end{aligned}
\]
where $a_1$ and $a_2$ denote contention intensity coefficients for scheduling and execution, respectively, and $t_{base1}$,\ \ $t_{base2}$ represent baseline times under single-thread execution. These parameters are determined by hardware platform and application context.

Combining the above, the total execution time of a single simulation cycle can be expressed as:
\begin{equation}
 T\left(k\right)=\left\{\begin{matrix}\left\lceil\frac{N}{k}\right\rceil\ast\frac{1}{\mu_e}+N\ast\left(\frac{k}{\mu_b\left(1-P_0\right)}-\frac{1}{k\mu_e}\right), \text{if } k\mu_e\le\mu_b\\
 \frac{N}{\mu_b}+\frac{1}{\mu_e}, \text{if } k\mu_e>\mu_b.\ \\\end{matrix}\right.
\end{equation}
The above equation shows that execution time is a piecewise-continuous, non-smooth function of the WT count $k$. Numerical analysis reveals that the optimal configuration occurs at $k^*\approx\mu_b / \mu_e$, indicating optimal performance when scheduler service rate balances with task execution rate.

\paragraph{PINO Predictor.}
While $\mu_e$ can be estimated online from probe-collected execution times, directly measuring $\mu_b$ is difficult due to microsecond-scale sensitivity and measurement overhead. We therefore introduce a Physics-Informed Neural Operator (PINO) to fuse the above model with historical traces for fast online prediction.
 

We define a neural operator $\mathcal{G}_\theta:a\rightarrow k^*=\mathcal{G}_\theta\left(a\right)$, which maps system state parameters $a$ to the optimal WT number $k^*$. We construct the feature vector $a=\{ N, \mu_e, k_{obs}, usr\_cpu, system\_cpu, VSZ, cswch\}$ from both application and hardware features. Although virtual memory usage (VSZ) has limited  impact on thread decisions, excessive memory consumption may trigger resource contention or even abnormal program termination; thus, it is included as an input to enhance robustness.

As illustrated in Figure~\ref{fig:AutoThread}, PINO predictor employs a multi-output head cross-attention architecture. Encoders separately process application and hardware features, cross-attention captures their deep interactions, and FC layers decode fused features into optimal WT count $k^*$ and contention coefficients $a_1$,\ $b_1$.
Physical loss:
\[
\mathcal{L}_{phy}=\Vert T(N,k,\mu_b,\mu_e)-T_{obs} \Vert_2+\alpha \Vert \frac{\mu_b}{\mu_e} -k^* \Vert_2
\]
Supervised loss and Regularization term:
\[
\mathcal{L}_{sup} = \Vert k^* - k_{obs} \Vert_2, \quad
\mathcal{L}_{reg} = \gamma \Vert \theta \Vert_2
\]
where $T$ represents the theoretical performance model (Equation 1), $\mu_b$ is derived from the contention model as $\mu_b=a_1k+b_1$, and $T_{obs}$ and $k_{obs}$\ denote the observed task queue execution time and iteratively measured optimal WT number, respectively. The weight $\alpha$  normalizes contributions of the two physical components. The overall training objective is 
$\mathcal{L}_{loss}=\mathcal{L}_{phy}+\lambda\mathcal{L}_{sup}+\mathcal{L}_{reg}$.
We incorporate Residual-Based Attention (RBA) \cite{anagnostopoulos2024residual} to dynamically adjust $\lambda$ between physical and supervised losses. In out-of-distribution scenarios, RBA automatically increases the physical-loss weight, strengthening reliance on the performance model and improving prediction robustness.

\subsection{Adaptive Dynamic Tuner}
Based on the trained PINO predictor, we design an adaptive dynamic tuner adopting a three-tier progressive adjustment strategy. The tuner aims to establish a closed-loop control process of prediction–execution–feedback–readjustment, so as to mitigate adverse effects caused by prediction errors, system noise, and measurement bias. The complete workflow is presented in Algorithm \ref{alg:autothread}.

\paragraph{Load-Aware Fast Prediction.}During each execution cycle, the system automatically records the average service time of the current task queue ${t}{cur}$ using probe pointer and computes the average processing time of the queue ${T}{cur}$ (lines 1-3). When a change in ${t}_{cur}$ exceeding a preset threshold is detected, the tuner immediately invokes PINO to predict the optimal WT count and update it (lines 4–6). This mechanism ensures fast responsiveness to sudden load changes.

\paragraph{CPU-Aware Periodic Adjustment.}To avoid continuous prediction overhead and handle gradually varying loads, a periodic checking mechanism is introduced. The check interval is dynamically adaptive: if prediction was triggered in the previous cycle, a short interval $C_1$ is used to verify and correct predictions quickly; otherwise, a longer interval $C_2$ is adopted to reduce monitoring overhead (line 7). At a checkpoint, if the observed $system\_cpu$ exceeds the  $ usr\_cpu$, the algorithm proactively reduces the number of WTs to mitigate contention (lines 8–16). This mechanism effectively maps CPU indicators of resource contention into direct signals for thread adjustment, thereby preventing performance degradation due to oversubscription.

\paragraph{Throughput-Driven Fine-tuning.}When the system is detected to be in a high-contention state, the tuner switches to a fine-tuning phase. In this phase, throughput of the task queue is directly optimized as the adjustment objective. If reducing threads leads to improved throughput, the adjustment is accepted and further attempts are made; otherwise, the change is rolled back and the phase is terminated (lines 17–25). Considering that undersubscription has a less detrimental effect on performance than the severe contention caused by oversubscription, the fine-tuning phase only considers reducing threads. Moreover, to avoid erroneous adjustments due to random fluctuations or transient anomalies, the tuner exits this phase only if throughput fails to improve over three consecutive simulation steps or if the thread count reaches the minimum allowed value (line 20).

\begin{algorithm}[t]
\caption{Three-Level Progressive Thread Adjustment}
\label{alg:autothread}
\begin{algorithmic}[1]
\Statex \textbf{Input:} $\delta$ (load change threshold), $C_1$ (short period), $C_2$ (long period)
\Statex \textbf{Output:}  WT count $k$ for each simulation step

\While{simulation not finished}
    \State $t_{\text{cur}} \gets \textsc{ProbeTaskTime}()$
    \State $T_{\text{cur}} \gets \textsc{ExecuteCycle}()$

    \If{${|t_{\text{cur}}-t_{\text{prev}}|}  / ({t_{\text{prev}}+\epsilon}) > \delta$ \textbf{and} 
    $\textit{tuning}=0$}
        \State $k \gets \textsc{PINO-Predict}$
        \State $\textit{predicted} \gets \textsc{True}$

    \ElsIf{$\textit{predicted}$ \textbf{and} $(\textit{tick} \bmod C_1 = 0)$}
        \State $(system\_cpu, usr\_cpu) \gets \textsc{GetCpuMetrics}()$
        \If{$ usr\_cpu <  system\_cpu$}
            \State $k \gets k - 1$, $\textit{tuning} \gets 1$
        \EndIf
        \State $\textit{predicted} \gets \textsc{False}$

    \ElsIf{$\neg\textit{predicted}$ \textbf{and} $(\textit{tick} \bmod C_2 = 0)$}
        \If{$ usr\_cpu <  usr\_cpu$}
            \State $k \gets k - 1$, $\textit{tuning} \gets 1$
        \EndIf

    \ElsIf{$\textit{tuning} \ge 1$ \textbf{and} $k > k_{\min}$}
        \If{$T_{\text{cur}} < T_{\text{prev}}$} \Comment{Throughput improved}
            \State $k \gets k - 1$, $\textit{tuning} \gets 1$
        \ElsIf{$\textit{tuning} < 3$} \Comment{Tolerate fluctuation}
            \State $\textit{tuning} \gets \textit{tuning} + 1$
        \Else \Comment{Rollback and exit tuning}
            \State $k \gets k + 1$, $\textit{tuning} \gets 0$
        \EndIf
    \EndIf

    \State $t_{\text{prev}} \gets t_{\text{cur}}$, $T_{\text{prev}} \gets T_{\text{cur}}$
    \State \textsc{Synchronize}(); \ \textsc{NextTick}()
\EndWhile
\end{algorithmic}
\end{algorithm}

\section{Evaluation}
In this section, we first introduce the experimental setup and current thread tuning methods. Subsequently, we evaluate the proposed method against these baselines in terms of system acceleration performance and adaptability. Finally, we conduct ablation studies to quantify the contributions of each optimization module and verify the robustness of our method with respect to various hyperparameters.
\subsection{Setup}
The training set is derived from averaged statistics over the entire execution under constant-load scenarios, where the execution time of each task remains unchanged throughout the simulation horizon. We split the dataset into training/testing/validation sets with a ratio of 0.7/0.2/0.1 to offline train different predictors. During training, each feature vector is taken as the average over the whole execution; since the load is constant, the optimal thread count label for each sample can be obtained by sweeping different thread counts and recording the total execution time.

In the testing phase, we use dynamic workloads to evaluate the speedup and adaptivity of each method. Specifically, within each simulation step, task execution times follow a Poisson distribution parameterized by $t$, while the parameter $t$ varies across steps as a square wave with frequency $roh$ (switching among [$t,1.5t,2t$]). Each method adjusts based on instantaneous values of the feature variables at a single time point, so as to emulate a practical deployment setting with offline training and online testing.

Experiments are conducted on two multicore architectures: AMD EPYC 9734 (112 Cores/224 Threads) and Intel Xeon Gold 6338 (32 Cores/64 Threads). The baselines include  \textbf{ADAPT-T}\cite{costa2019adapt} based on runtime feedback control; \textbf{Otter}\cite{luan2022online} employing search strategies; \textbf{Thread Reinforcer}\cite{pusukuri2011thread} utilizing system resource utilization; \textbf{Xgboost}\cite{akash2021machine} as an advanced representative of machine learning-based prediction. Additionally, a fixed thread count configuration was used as a control experiment, denoted as \textbf{Static}. All methods were implemented on the Repast HPC system\cite{zhu2017hierarchical}. Each test case was executed at least five times to obtain optimal measurement data.


\begin{figure}[t]
    \centering
    \includegraphics[width=\linewidth]{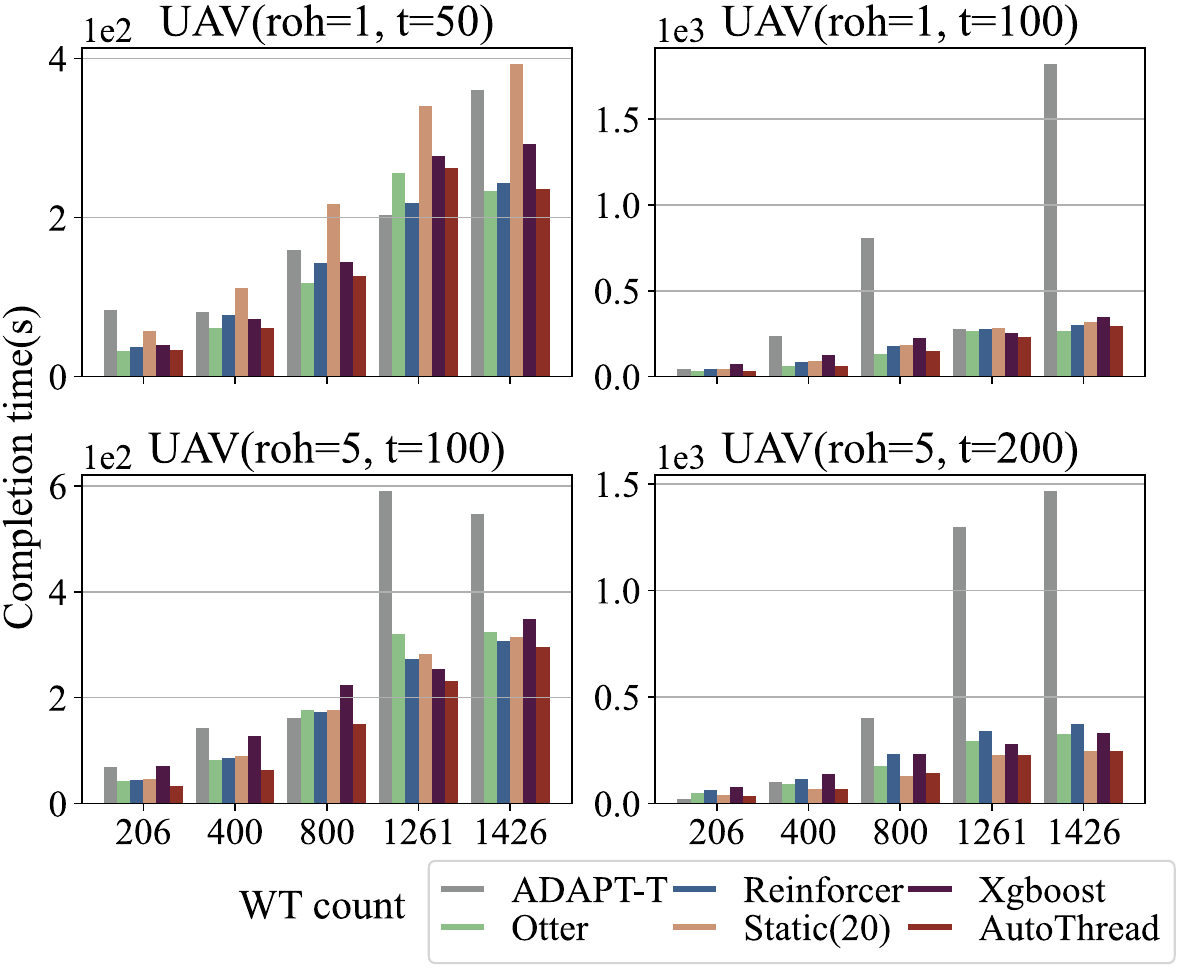}
    \caption{Runtime comparison of different methods on AMD platform for UAV}
    \label{fig:figure9}
\end{figure}

\begin{figure}[t]
    \centering
    \includegraphics[width=1\linewidth]{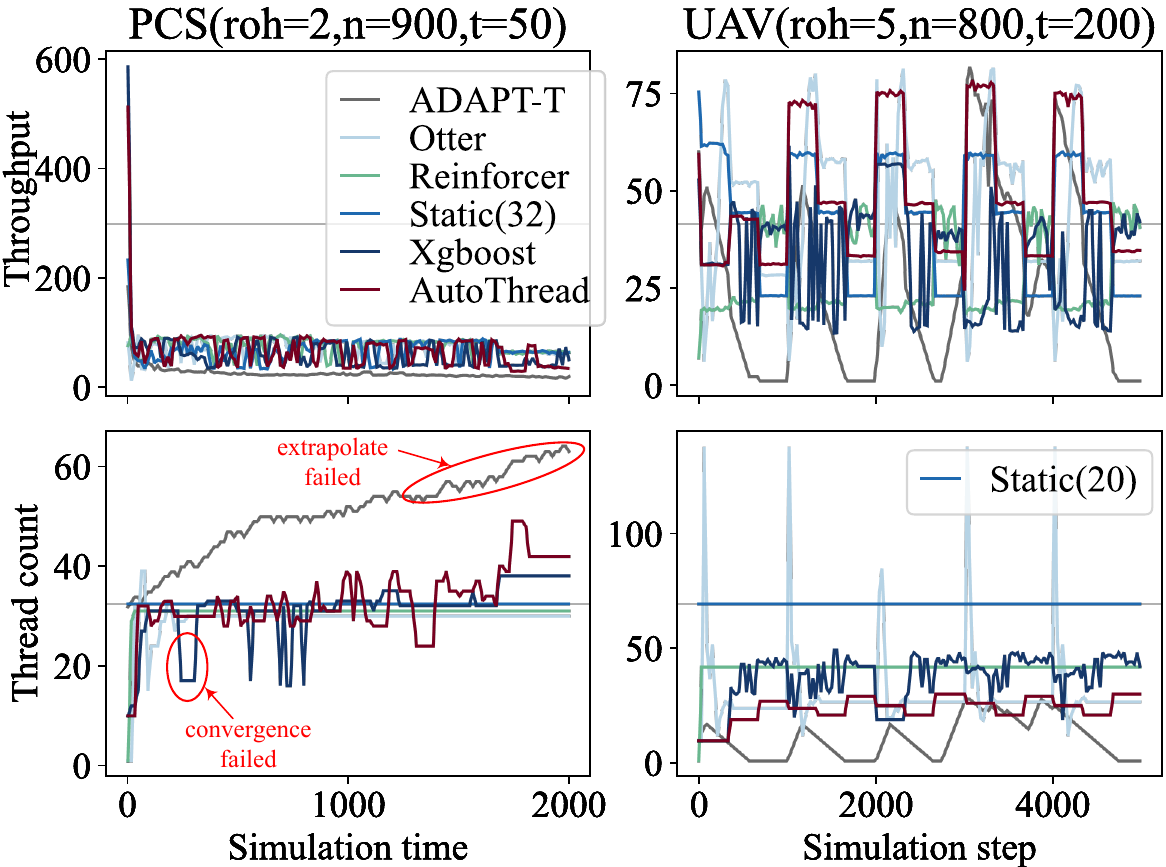}
    \caption{Adaptiveness comparison on PCS and UAV}
    \label{fig:figure10}
\end{figure}

\begin{figure}[t]
    \centering
    \includegraphics[width=1\linewidth]{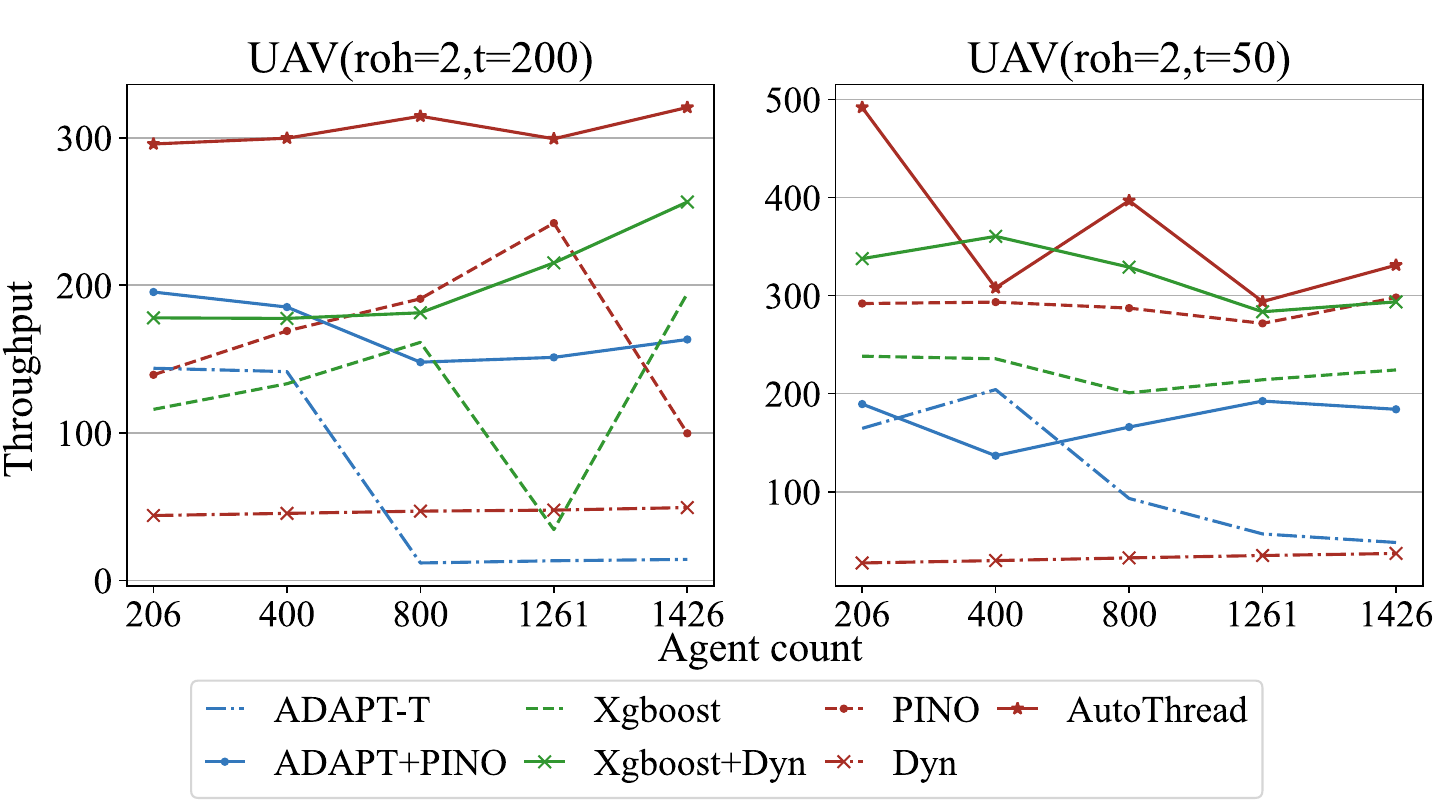}
    \caption{Results of ablation experiments}
    \label{fig:figure11}
\end{figure}

\subsection{Performance Experiments}
We run the UAV benchmark on the AMD platform with the execution horizon set to 5000 and the load threshold $\delta= 0.3$, and record the total runtime from end to-end under different thread adjustment schemes; the results are shown in Fig. \ref{fig:figure9}. Compared to ADAPT-T, AutoThread saves up to 83.8\% of runtime in different scenarios. Over 20 test scenarios, AutoThread achieves an average runtime of 149.32 s and delivers 10.3\% and 17.2\% higher speedup than the state-of-the-art Otter and Reinforcer, respectively. Notably, we set the thread count of the fixed strategy to 20, i.e., the average output of all tuning schemes. Even when accounting for the additional overhead introduced by frequent thread adjustments, AutoThread still attains, on average, an 18.4\% higher speedup than the fixed strategy with prior knowledge, with a maximum performance improvement of 62.6\%.

\subsection{Adaptiveness Experiments}
We run the PCS and UAV applications on both Intel and AMD platforms under different load-variation intensities (frequency $roh$, amplitude $t$) , and record the system throughput over environment steps to evaluate the adaptability of different schemes, where throughput denotes the number of tasks processed per millisecond. Figure~\ref{fig:figure10} shows that, in practical scenarios, when competing methods successively exhibit non-convergent oscillations and extrapolation failures, AutoThread still maintains strong parallel speedup. On the PCS case, AutoThread attains an average throughput of 536.83, improving over ADTPT-T and Xgboost by 37.7\% and 2.3\%, respectively, while remaining below Reinforcer’s 718.55. Under the UAV case with larger workload fluctuations, AutoThread’s average throughput is 1.8× and 1.7× that of Reinforcer and Xgboost, respectively.

\subsection{Ablation Experiments}
We ablate the PINO predictor and the Adaptive Dynamic Tuner to investigate the performance contribution of each component. The compared variants include: using only the physics-informed neural operator network prediction (\textbf{PINO}); using the dynamic tuner with the maximum thread count (\textbf{Dyn}); applying the Dyn tuner on top of the Xgboost output (\textbf{Xgboost+Dyn}); and pairing the ADAPT-T tuner with the output of PINO (\textbf{ADAPT+PINO}). Figure~\ref{fig:figure11} reports the average throughput of different schemes when running the UAV application on the ADM platform. The results show that both the PINO predictor and the Adaptive Dynamic Tuner effectively improve runtime performance. Across 10 test scenarios, the Dyn technique increases the task processing speed of Xgboost by 87.8\% on average, with a maximum improvement of 55.2×. PINO increases the task processing speed of the ADAPT-T tuner by 3.9× on average. Moreover, AutoThread improves average throughput by 1.3× and 2.0× over Xgboost+Dyn and ADAPT+PINO, respectively, demonstrating the combined benefits of the two techniques.

\subsection{Robustness Experiment}
\begin{figure}
    \centering
    \includegraphics[width=\linewidth]{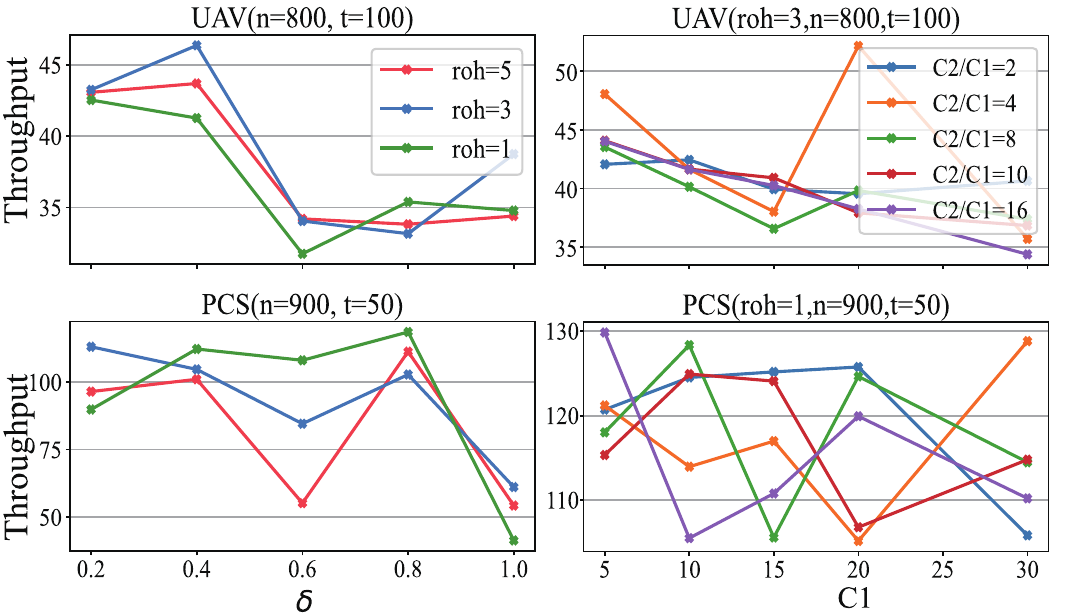}
    \caption{Robustness analysis of $\delta$ and $C_1/C_2$ parameter}
    \label{fig:figure12}
\end{figure}

We further examine the impact of parameters $\delta$, $C_1$ and $C_2$ on AutoThread’s performance. The parameter
$\delta$ represents the maximum tolerable degree of workload change for thread adjustment. An overly small value leads to frequent thread changes due to random hardware jitter, whereas an overly large value prevents timely adaptation to workload changes. AutoThread’s robustness to $\delta$ is shown in Figure~\ref{fig:figure12}. Benefiting from the theoretical constraints of the physical model and the error-correction capability of the dynamic tuner, AutoThread provides stable speedup even under workload inputs with substantial randomness. In contrast, an excessively large $\delta$ weakens the predictor’s sensitivity to workload changes, resulting in performance degradation.

Figure~\ref{fig:figure12} shows the impact of different $C_1/C_2$ combinations on system performance. We find that a smaller $C_1$ enables faster response; however, an overly small $C_2$ causes frequent periodic checks and increases system overhead, while an overly large $C_2$ may cause the system to miss critical tuning opportunities. The optimal ratio is approximately 2–4, within which short-cycle validation and long-cycle monitoring achieve a good balance—responding quickly to workload changes while avoiding excessive monitoring overhead.

\subsection{Discussion}
The experimental results clearly demonstrate the advantages of AutoThread in sustaining parallel speedup and substantially reducing execution time, particularly in environments with fluctuating workloads. The key strength of AutoThread lies in the deep integration of data-driven representational capacity with the structural constraints imposed by physical models: on the one hand, it learns latent prior patterns from runtime data to enable rapid prediction under dynamic workloads; on the other hand, by introducing interpretable theoretical models as inductive biases and training constraints, it anchors the prediction process in consistent mechanistic logic, thereby significantly improving robustness and the achievable performance upper bound while preserving interpretability and transferability. This form of mechanism-guided learning coupled with data-corrected inference is central to AutoThread’s superior effectiveness.

Specifically, first, guided by a performance model, PINO enforces queueing-theoretic constraints on the neural network’s predictive outputs, endowing the predictions with strong generalization and physical plausibility; consequently, it can provide reliable initial estimates of thread counts even under workload conditions whose distributions differ from those seen during training. Second, the dynamic tuning mechanism employs a three-level strategy to perform online corrections to the predicted results, while accounting for the overhead of monitoring and adjustment. This enables the system to rapidly converge to a near-optimal thread configuration when the workload changes, thereby accelerating the simulation environment’s runtime efficiency.

\section{Related Work}

Many WT tuning methods have been proposed across domains; however, compared with this study, they insufficiently address dynamic workload fluctuations or generalizability. Static modeling methods set fixed thread counts through empirical heuristics or offline analysis, such as using a multiple of physical cores\cite{montesano2024spatial} or estimating optimum from the ratio of thread creation to maintenance overhead \cite{luoextensible}. Although simple and efficient, these strategies struggle in complex, volatile environments because they cannot adjust to workload fluctuations in real time.

Machine learning has enabled data-driven thread-configuration prediction, including Reinforcement Learning\cite{xu2012url}, Ensemble Learning \cite{chao2025memory}, and online search optimization\cite{stetsenko2019thread}, with applications in large-scale cluster management. However, these methods often rely on high-quality training data; under dynamic scenarios absent from training, pure data-driven models may generalize poorly. Another common category uses adaptive regulation based on online feedback control \cite{albalawi2025dynamic}. These methods incrementally adjust WT count, one thread at a time, using metrics such as request arrival rates and service times until performance no longer improves. Although practical for cloud services, such feedback-driven regulation often suffers from slow convergence, latency, and performance oscillations in large thread spaces or under frequent workload variations \cite{bahadur2025poolrunner}.

\section{Conclusion}
This paper proposes AutoThread, a thread-count tuning method for ABS systems to accelerate RL inference. The core contributions are as follows: (1) we open-source a multithreaded trajectory dataset collected from a DES simulation environment, and empirically derive three key insights; (2) we design a PINO-based thread predictor and establish a theoretical model relating the number of WTs to runtime, which is used to constrain and guide model training; and (3) we develop an efficient hybrid thread adjustment method that incorporates a load-aware adaptive tuning strategy on top of PINO’s predictions. Experimental results show that AutoThread delivers excellent parallel speedup and strong workload adaptivity. The proposed approach is portable and can be applied to accelerate other systems based on a shared event queue. Future work will explore batch-sample thread tuning methods for accelerating RL training.

\section*{Acknowledgments}

We thank the anonymous IJCAI reviewers for their insightful comments. This work was partially supported by the National Natural Science Foundation of China (NSFC) under Grant No.62503491.

\section*{Contribution Statement}
Jiming Su and Hantao Hua contributed equally to this work. Feng Zhu is the corresponding author.

\bibliographystyle{named}
\bibliography{ijcai26}

@article{montesano2024spatial, title={Spatial/temporal locality-based load-sharing in speculative discrete event simulation on multi-core machines}, author={Montesano, Federica and Marotta, Romolo and Quaglia, Francesco}, journal={ACM Transactions on Modeling and Computer Simulation}, volume={35}, number={1}, pages={1--31}, year={2024}, publisher={ACM New York, NY} }

@inproceedings{costa2019adapt, title={Adapt-t: An adaptive algorithm for auto-tuning worker thread pool size in application servers}, author={Costa, Nilushan and Jayasinghe, Malith and Atukorale, Ajantha and Abeysinghe, Supun and Perera, Srinath and Perera, Isuru}, booktitle={2019 IEEE Symposium on Computers and Communications (ISCC)}, pages={1--6}, year={2019}, organization={IEEE} }

@inproceedings{pusukuri2011thread, title={Thread reinforcer: Dynamically determining number of threads via os level monitoring}, author={Pusukuri, Kishore Kumar and Gupta, Rajiv and Bhuyan, Laxmi N}, booktitle={2011 IEEE International Symposium on Workload Characterization (IISWC)}, pages={116--125}, year={2011}, organization={IEEE} }

@inproceedings{akash2021machine, title={Machine Learning Based Thread Pool Tuning via Program Analysis}, author={Akash, Lakindu and Fernando, Duneesha and Jayasinghe, Malith and Keppitiyagama, Chamath and Thangarajah, Kishanthan}, booktitle={2021 IEEE 23rd Int Conf on High Performance Computing \& Communications; 7th Int Conf on Data Science \& Systems; 19th Int Conf on Smart City; 7th Int Conf on Dependability in Sensor, Cloud \& Big Data Systems \& Application (HPCC/DSS/SmartCity/DependSys)}, pages={648--653}, year={2021}, organization={IEEE} }

@article{xu2012url, title={URL: A unified reinforcement learning approach for autonomic cloud management}, author={Xu, Cheng-Zhong and Rao, Jia and Bu, Xiangping}, journal={Journal of Parallel and Distributed Computing}, volume={72}, number={2}, pages={95--105}, year={2012}, publisher={Elsevier} }

@article{carothers2002ross, title={ROSS: A high-performance, low-memory, modular Time Warp system}, author={Carothers, Christopher D and Bauer, David and Pearce, Shawn}, journal={Journal of parallel and distributed computing}, volume={62}, number={11}, pages={1648--1669}, year={2002}, publisher={Elsevier} }

@inproceedings{stetsenko2019thread, title={Thread pool parameters tuning using simulation}, author={Stetsenko, Inna V and Dyfuchyna, Oleksandra}, booktitle={International Conference on Computer Science, Engineering and Education Applications}, pages={78--89}, year={2019}, organization={Springer} }

@article{bahadur2025poolrunner, title={PoolRunner: An Extensible Performance Testing Simulation Tool for Thread-Pool Middleware}, author={Bahadur, Faisal and Ahmad, Zulfiqar and Algarni, Abdulmohsen}, journal={IEEE Access}, year={2025}, publisher={IEEE} }

@article{anagnostopoulos2024residual, title={Residual-based attention in physics-informed neural networks}, author={Anagnostopoulos, Sokratis J and Toscano, Juan Diego and Stergiopulos, Nikolaos and Karniadakis, George Em}, journal={Computer Methods in Applied Mechanics and Engineering}, volume={421}, pages={116805}, year={2024}, publisher={Elsevier} }

@article{zhu2017hierarchical,
  title={A hierarchical composite framework of parallel discrete event simulation for modelling complex adaptive systems},
  author={Zhu, Feng and Yao, Yiping and Tang, Wenjie and Tang, Jun},
  journal={Simulation Modelling Practice and Theory},
  volume={77},
  pages={141--156},
  year={2017},
  publisher={Elsevier}
}

@misc{flightaware_uk_2025_online,
  author   = {{FlightAware}},
  title    = {FlightAware - Flight Tracker / Flight Status},
  year     = {2025},
  howpublished = {\url{https://uk.flightaware.com/}},
  note     = {Accessed: 2025-10-25},
  organization = {FlightAware},
  address  = {Eleven Greenway Plaza, Suite 2900, Houston, Texas 77046, USA}
}

@article{feriani2021single, title={Single and multi-agent deep reinforcement learning for AI-enabled wireless networks: A tutorial}, author={Feriani, Amal and Hossain, Ekram}, journal={IEEE Communications Surveys \& Tutorials}, volume={23}, number={2}, pages={1226--1252}, year={2021}, publisher={IEEE} }

@article{hu2024review, title={A review of research on reinforcement learning algorithms for multi-agents}, author={Hu, Kai and Li, Mingyang and Song, Zhiqiang and Xu, Keer and Xia, Qingfeng and Sun, Ning and Zhou, Peng and Xia, Min}, journal={Neurocomputing}, volume={599}, pages={128068}, year={2024}, publisher={Elsevier} }

@article{wu2022mixed, title={Mixed opinion dynamics based on DeGroot model and Hegselmann--Krause model in social networks}, author={Wu, Zhibin and Zhou, Qinyue and Dong, Yucheng and Xu, Jiuping and Altalhi, Abdulrahman H and Herrera, Francisco}, journal={IEEE Transactions on Systems, Man, and Cybernetics: Systems}, volume={53}, number={1}, pages={296--308}, year={2022}, publisher={IEEE} }

@article{eldeeb2025offline, title={Offline and distributional reinforcement learning for wireless communications}, author={Eldeeb, Eslam and Alves, Hirley}, journal={arXiv preprint arXiv:2504.03804}, year={2025} }

@article{riboldi2025formation, title={Formation Flight of Fixed-Wing UAVs: Dynamic Modeling, Guidance Design, and Testing in Realistic Scenarios}, author={Riboldi, Carlo ED and Tomasoni, Marco and others}, journal={Aerospace}, volume={12}, number={3}, pages={1--49}, year={2025} }

@article{morgado2025evaluating, title={Evaluating end-to-end autonomous driving architectures: a proximal policy optimization approach in simulated environments}, author={Morgado, {\^A}ngelo and Ota, Kaoru and Dong, Mianxiong and Pombo, Nuno}, journal={Autonomous Intelligent Systems}, volume={5}, number={1}, pages={14}, year={2025}, publisher={Springer} }

@inproceedings{belsare2022reinforcement, title={Reinforcement learning with discrete event simulation: the premise, reality, and promise}, author={Belsare, Sahil and Badilla, Emily Diaz and Dehghanimohammadabadi, Mohammad}, booktitle={2022 Winter Simulation Conference (WSC)}, pages={2724--2735}, year={2022}, organization={IEEE} }

@article{risco2022unified, title={A unified cloud-enabled discrete event parallel and distributed simulation architecture}, author={Risco-Mart{\'\i}n, Jos{\'e} L and Henares, Kevin and Mittal, Saurabh and Almendras, Luis F and Olcoz, Katzalin}, journal={Simulation Modelling Practice and Theory}, volume={118}, pages={102539}, year={2022}, publisher={Elsevier} }

@article{samadi2024time, title={Time-predictable task-to-thread mapping in multi-core processors}, author={Samadi, Mohammad and Royuela, Sara and Pinho, Luis Miguel and Carvalho, Tiago and Qui{\~n}ones, Eduardo}, journal={Journal of Systems Architecture}, volume={148}, pages={103068}, year={2024}, publisher={Elsevier} }

@article{li2025adwtune, title={ADWTune: an adaptive dynamic workload tuning system with deep reinforcement learning}, author={Li, Cuixia and Wang, Junhai and Shi, Jiahao and Liu, Liqiang and Zhang, Shuyan}, journal={Complex \& Intelligent Systems}, volume={11}, number={4}, pages={192}, year={2025}, publisher={Springer} }

@article{chao2025memory, title={Memory-efficient and adaptive heterogeneous framework for gate-level fault simulation}, author={Chao, Zhiteng and Gu, Feng and Huang, Junying and Li, Wenjie and Ye, Jing and Li, Huawei and Li, Xiaowei}, journal={ACM Transactions on Design Automation of Electronic Systems}, volume={30}, number={5}, pages={1--27}, year={2025}, publisher={ACM New York, NY} }

@article{luan2022online, title={Online thread auto-tuning for performance improvement and resource saving}, author={Luan, Guangqiang and Pang, Pu and Chen, Quan and Xue, Shuai and Song, Zhuo and Guo, Minyi}, journal={IEEE Transactions on Parallel and Distributed Systems}, volume={33}, number={12}, pages={3746--3759}, year={2022}, publisher={IEEE} }

@article{albalawi2025dynamic, title={Dynamic scheduling strategies for cloud-based load balancing in parallel and distributed systems}, author={Albalawi, Nasser S}, journal={Journal of Cloud Computing}, volume={14}, number={1}, pages={33}, year={2025}, publisher={Springer} }

@article{jancauskas2019predicting,
  title={Predicting queue wait time probabilities for multi-scale computing},
  author={Jancauskas, Vytautas and Piontek, Tomasz and Kopta, Piotr and Bosak, Bartosz},
  journal={Philosophical Transactions of the Royal Society A},
  volume={377},
  number={2142},
  pages={20180151},
  year={2019},
  publisher={The Royal Society Publishing}
}

@article{lin2025hap, title={HAP: Hybrid Adaptive Parallelism for Efficient Mixture-of-Experts Inference}, author={Lin, Haoran and Yu, Xianzhi and Zhao, Kang and Bao, Han and Zhan, Zongyuan and Hu, Ting and Liu, Wulong and Yin, Zekun and Li, Xin and Liu, Weiguo}, journal={arXiv preprint arXiv:2508.19373}, year={2025} }

@inproceedings{zhang2021parallel, title={Parallel actors and learners: A framework for generating scalable RL implementations}, author={Zhang, Chi and Kuppannagari, Sanmukh Rao and Prasanna, Viktor K}, booktitle={2021 IEEE 28th International Conference on High Performance Computing, Data, and Analytics (HiPC)}, pages={1--10}, year={2021}, organization={IEEE} }

@article{luoextensible,
author = {Luo, Xiaoxuan and Lin, Weiwei and Li, Jiachun and Chen, Fan and Zhong, Haocheng and Li, Keqin},
title = {An Extensible Thread Throttling Method for Multiple OpenMP Parallel Programs},
year = {2025},
publisher = {Association for Computing Machinery},
address = {New York, NY, USA},
issn = {1539-9087},
url = {https://doi.org/10.1145/3769679},
doi = {10.1145/3769679},
journal = {ACM Trans. Embed. Comput. Syst.},
month = sep
}

\end{document}